\documentclass[10pt,twocolumn]{article}

\usepackage[square,numbers]{natbib}
\usepackage{cvpr}
\usepackage{xcolor,xspace}
\usepackage{algorithm}
\usepackage[noend]{algpseudocode}
\usepackage{amsmath}
\usepackage{amssymb}
\usepackage{booktabs}
\usepackage{balance}
\usepackage{float}
\usepackage{amsfonts}      
\usepackage{url}

\newcommand{\comment}[1]{}

\usepackage{times}
\usepackage{epsfig}
\usepackage{graphicx}

\usepackage{subfigure}
\usepackage{listings}

\usepackage[pagebackref=true,breaklinks=true,colorlinks,bookmarks=false,]{hyperref}

\cvprfinalcopy 

\def\cvprPaperID{XXXX} 

\definecolor{light-gray}{gray}{0.95}

\ifcvprfinal\fi
\begin{document}

\title{Leveraging Speculative Sampling and KV-Cache Optimizations Together for Generative AI using OpenVINO}
\author{
Haim Barad
\and
Ekaterina Aidova
\and
Yury Gorbachev
}

\maketitle

\thispagestyle{empty}

\begin{abstract}

Inference optimizations are critical for improving user experience and reducing infrastructure costs and power consumption. In this article, we illustrate a form of dynamic execution known as speculative sampling to reduce the overall latency of text generation and compare it with standard autoregressive sampling. This can be used together with model-based optimizations (e.g. quantization) to provide an optimized solution. Both sampling methods make use of KV caching. A Jupyter notebook and some sample executions are provided.

\end{abstract}

{\let\thefootnote\relax\footnote{{Published in the \textit{OpenVINO Blog}, openvino.ai 2023}}}

\section{Introduction}

As model sizes grow, Generative AI implementations require significant inference resources. This not only increases the cost per generation from a prompt, but also increases the power consumption used to serve such requests.

Inference optimizations for text generation are essential for reducing latency, infrastructure costs, and power consumption. This can lead to an improved user experience and increased efficiency in text generation tasks.

Another necessary condition is that the optimizations are compatible with each other. That is, implementing a certain optimization should not preclude or conflict with other optimizations. There are several levels of optimizations that can provide significant speedup without "bumping into each other" in a way that will compromise overall efficiency.

We provide code to illustrate speculative sampling with KV caching. This is available at \href{https://github.com/openvinotoolkit/openvino_notebooks/tree/latest/notebooks/speculative-sampling}{speculative sampling}.

\section{What is the problem?}
Let us take a look at one such collection of optimizations that would be desired for many text generation applications. We will consider the following optimizations given a full model:
\begin{itemize}
    \item \textbf{Model-Based Optimizations} - often we don't expect problems to arise if quantization (a commonly used model-based optimization) was tested by itself. However, there could be some unexpected effects from the quantization and sampling of models. A full discussion on quantization is beyond the scope of this article (but is assumed to be essential in high-performance inference); we will focus on dynamic execution methods in this work.
    \item \textbf{KV Caching} (or Past-Value Caching) - autoregressive sampling predicts the next value in a series of tokens. Computations are performed on these tokens to create the prediction of the next token. The expanded collection of tokens (all previously appended with the newly generated token from the previous pass) now goes through another pass, and this continues until the number of tokens requested is reached. To avoid a lot of repetitive calculations on the past tokens, the intermediate values are stored in a KV cache \cite{pope_efficiently_2022}. This method is very standard (enabled in HuggingFace by default) and poses no risk to accuracy. The only downside is that the KV cache can be quite large and increase memory requirements for the autoregressive process.
    \item \textbf{Speculative Sampling} - A form of dynamic execution, there has been a lot of published research in this area about using a smaller, draft model to produce samples that should be "good enough" much of the time, and occasionally reject candidates and pay the price of the full model when needed. This method has been published, with slight differences, in several independent research publications. \cite{schuster_confident_2022}\cite{belrose_eliciting_2023}\cite{chen_accelerating_2023}\cite{kim_big_2023}\cite{gante_joao_assisted_nodate}\cite{stern_blockwise_2018}
\end{itemize}

In order to gain an appreciation of why speculative sampling works, let us take a step back and visit Autoregressive Sampling, combined with KV caching. We will see that the process is memory-bound allowing us to essentially test K tokens on the target model, in parallel, for the same cost as sampling just one token. So having a decent acceptance rate means that many of the tokens are generated fast enough to compensate for the extra overhead of generating on a draft model and then checking the batch of K candidate tokens in a target model.

\section{Autoregressive vs Speculative Sampling}
A method of text generation is to generate next tokens based upon a probability conditioned on previous tokens, as given by:
\begin{equation}
    p(\Tilde{x}_{n+1} | x_1, ..., x_n)
\end{equation}
This is known as autoregressive sampling \cite{lai_understanding_2017} and is now a standard method of text-generation in generative models. This could be followed by one of several methods to select the token at $n+1$, for example, argmax or randomly selected from top-p. 

Note that sampling of models is memory intensive. Shazeer \cite{shazeer_fast_2019} shows that the ratio of memory access to arithmetic operations is very memory intensive for transformer-based sequential sampling. Chen et al. \cite{chen_accelerating_2023} attribute the overall sampling time for large transformer-based models to linear layers, attention, and collective operations (all-reduce). We focus on a batch size of one for inference, but we can leverage a batch size of K words (sampled from a smaller draft model) to be evaluated in the target model together, taking about the same time as sampling a single token from the target model. For a reasonable value of K, we can, therefore, leverage the smaller draft model for much of the text generation, using the target model less often for evaluation (i.e., acceptance or rejection) and single token generation when rejection occurs. We have seen a significant increase in throughput using this method.

However, the draft model and target model have different sizes that would be represented in a KV cache, so the challenge is to take advantage of separate optimization strategies simultaneously. For this article, we assume a quantized model and leverage KV caching together with Speculative Sampling.

Note that the authors \cite{chen_accelerating_2023} prove that the target distribution is recovered when performing speculative sampling - this guarantees the same sampling quality as autoregressive sampling on the target itself. Therefore, the situations where not leveraging speculative sampling is not worthwhile have to do with the case where there are not enough savings in the relative size of the draft model or the acceptance rate of the draft model is not high enough to benefit from the smaller size of the draft model.

\section{Selected Optimizations}
Inference optimization is a process that combines model-based and execution-based optimizations to improve the performance of machine learning models. Model-based optimizations involve changes to the model architecture, such as pruning or quantization, while dynamic execution-based optimizations involve changes to the way the model is executed, such as conditional execution, path selection, and caching. When these two types of optimizations are combined, inference optimization can significantly improve the performance of machine learning models.

Depending on many factors, you might choose from a variety of optimizations. In this case, we choose the model-based optimizations provided by HuggingFace Optimum and OpenVINO, as well as two execution-based optimizations (KV caching and speculative sampling).

\subsection{OpenVINO Model Optimization with HuggingFace Optimum}
Optimum Intel can be used to load optimized models from the \href{https://huggingface.co/docs/optimum/intel/hf.co/models}{Hugging Face Hub} and create pipelines to run an inference with OpenVINO Runtime using Hugging Face APIs. The Optimum Inference models are API compatible with Hugging Face Transformers models.  This means that we just need to replace `AutoModelForXxx` class with the corresponding `OVModelForXxx` class.

In our case, we are using OVModelForCausalLM instead of AutoModelForCausalLM. For example, when we load a model $from\_pretrained$, we call it in a similar fashion:
\lstset{
  basicstyle=\ttfamily,
  columns=fullflexible,
  frame=single,
  breaklines=true,
  postbreak=\mbox{\textcolor{red}{$\hookrightarrow$}\space},
}
\begin{lstlisting}[language=Python]
model_id = "meta-llama/Llama-2-7b-chat-hf"
model = OVModelForCausalLM.from_pretrained(model_id)
\end{lstlisting}

The initialization of the model class starts by calling the $from\_pretrained$ method. When downloading and converting the Transformers model, the parameter $from\_transformers=True$ should be added. We can save the converted model for next use with the $save\_pretrained$ method.
The tokenizer class and pipelines API are compatible with Optimum models.

\subsection{KV Caching}
Also known as "Past Value Caching", a KV cache stores intermediate model values that are generated in previous computations \cite{pope_efficiently_2022}. If the text generation is using an autoregressive methodology, then subsequent iterations can leverage these calculations (thus, the name Past Value caching or Key Value caching). The cache can become quite large, and this is the trade-off of not having to repeat these calculations for future iterations.

\begin{figure}[h]
  \centering
  \includegraphics[width=\linewidth]{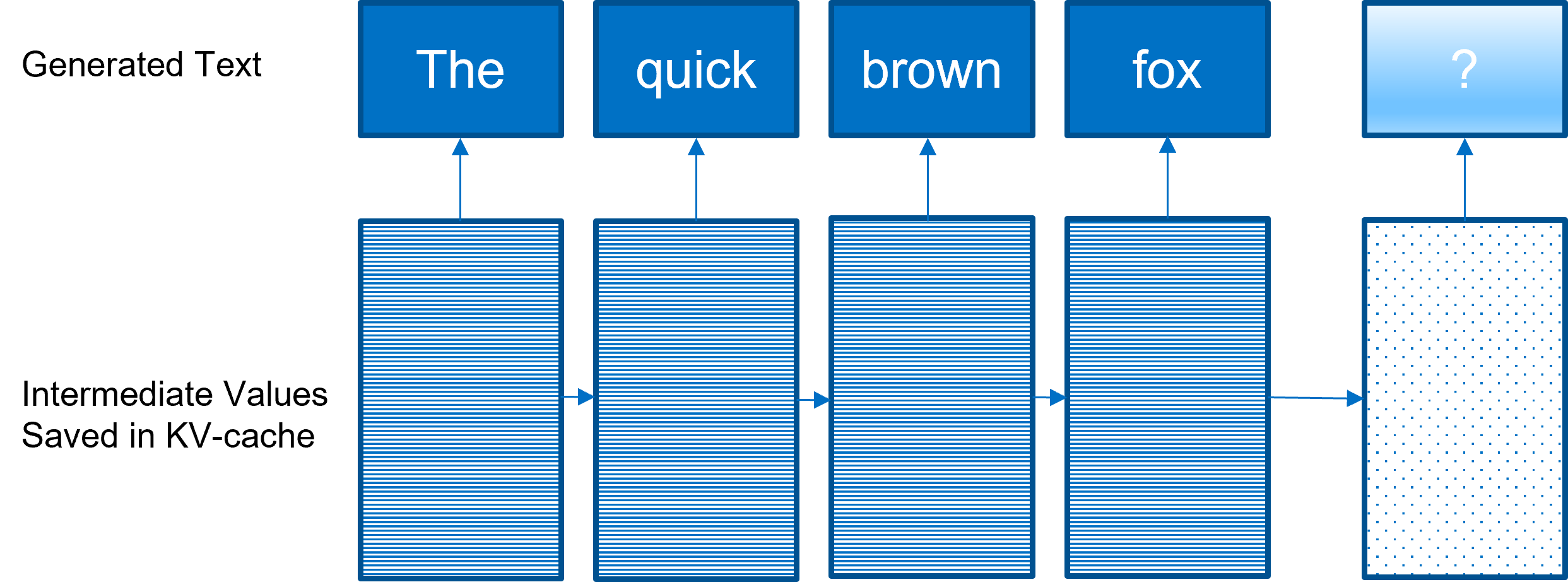}
  \caption{Autoregressive Sampling with KV caching}
\end{figure}
In Figure 1, the next value to generate after the word "fox" is dependent on all of the values generated previously to be fed into the attention mechanism for each of the past tokens. In order to save on repeated calculations, we can store these values in a cache (which can become quite large), and this provides a significant speed up in autoregressive text generation.

\subsection{Speculative Sampling}
Autoregressive text generation (even with optimizations such as KV cache) still requires a complete model for each token generated. Following in the footsteps of many Dynamic Execution methods (such as Early Exit), we can often provide sufficiently accurate answers with short cuts when the tasks being addressed are simple and accurate.

\begin{figure}
    \centering
    \includegraphics[width=0.9\linewidth]{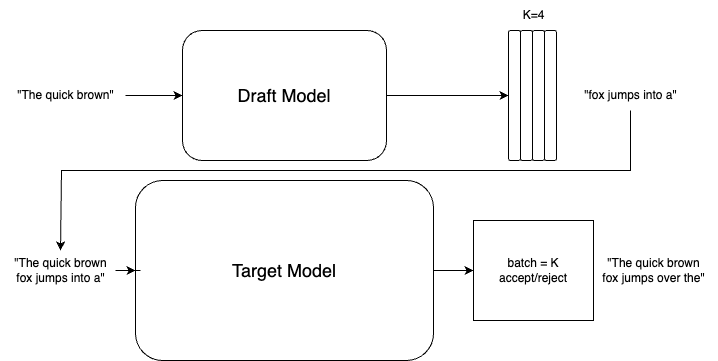}
    \caption{Speculative Sampling, K=4}
    \label{fig:speculative-sampling}
\end{figure}

Speculative Sampling \cite{chen_accelerating_2023} accelerates transformer decoding by using a smaller "draft model" for a short sequence of calls and uses the full-sized  "target model" to qualify and accept (or reject) the projected results of the draft model (see Figure \ref{fig:speculative-sampling} illustrated for K=4). Other approaches, such as BiLD \cite{kim_big_2023} and Assisted Generation \cite{gante_joao_assisted_nodate}, take a similar approach in that smaller models are pretty good at generating much of the text and validation (based on the full model) will occasionally reject tokens generated from the smaller model. As we will see, this will result in an overall speed-up.

\subsection{So, what's the problem?}
A KV cache is standard for situations like this. In fact, HuggingFace already enables such a cache by default. The primary reason to disable the cache in a generative scheme is only due to the memory requirements of such a cache.

\begin{figure}
    \centering
    \includegraphics[width=0.5\linewidth]{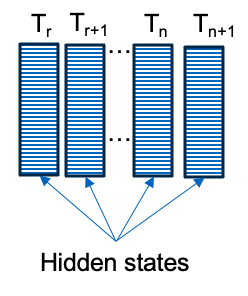}
    \caption{Hidden States in Target KV Cache}
    \label{fig:hidden-states}
\end{figure}

However, speculative sampling requires two separate models (target and draft), which both have different model sizes and, therefore, different sizes for the intermediate values that are stored in the KV cache. Furthermore, as we sample from the draft model, the target model's KV cache can become stale, but we still do not want to have redundant calculations. Do we have two separate caches? Is having a KV cache for the target model worth the memory footprint and the fact that its cache is only occasionally updated?

\section{Solution}

OpenVINO provides a collection of ready-to-run Jupyter notebooks to learn and experiment with the OpenVINOTM toolkit. This repository is available at \href{https://github.com/openvinotoolkit/openvino_notebooks}{Github}. We have provided a notebook with sample code to illustrate \href{https://github.com/openvinotoolkit/openvino_notebooks/tree/latest/notebooks/speculative-sampling}{speculative sampling}.

The speculative sampling notebook provides a clear example of how to take advantage of off-the-shelf models (e.g. HuggingFace), exploit OpenVINO classes, and leverage multiple sizes of a model for speculative sampling (and compare it with autoregressive sampling). By default, the notebook uses GPT2. If you have a machine with a lot of memory (at least 64GB), you might want to try the DollyV2 models. Other models can also be tried.

From our experience, it is best if the ratio of the sizes of the target model compared with the draft model is at least 10x. The reason is that the K candidate tokens generated from the draft model and the evaluation of the K tokens in the target model (batch size K) need to be less than autoregressively generating K tokens from the target model. This should provide a good speedup in terms of reduced latency.

\section{Experiments}

We ran the Dolly V2 model for a value of K=5 and did this for scenarios where the number of generated tokens was limited to 40 and 100, respectively. These results are shown in the Appendix.

\section{Conclusions}
In this paper, we are faced with a challenge to integrate KV Caching together with Speculative Sampling in a way that produces the most efficient method (combined with a quantized model).

{\small
\bibliographystyle{ieeetr}
\bibliography{references}

\begin{thebibliography}{1}
\bibitem{shazeer_fast_2019}
N.~Shazeer, ``Fast transformer decoding: One write-head is all you need.''
\end{thebibliography}
}

\section{Appendix}
\subsection{Samples using Databricks DollyV2}
The sampling was performed for a target model of size 12B and a draft model of size 3B for K=5 and for different values of N (number of tokens to generate)

\subsubsection{N=40}

Autoregressive Decode\\
---------------------\\
Time = 19.87s\\
Text = \textbf{Alan Turing theorized that computers would one day become} self-aware and try to take over the world. He was right, and it’s going to take a lot more than just a virus to stop them.
 tech.li
The
\\
\\
Speculative Decode\\
------------------\\
Time = 13.61s\\
Text = \textbf{Alan Turing theorized that computers would one day become} souringly advanced that they would have the ability to break the law I think that is a reasonable request I think that is a reasonable request I think that is a reasonable request I think that is a good\\
-------------------------------------------------------------------\\
\\
Autoregressive Decode\\
---------------------\\
Time = 18.45s\\
Text = \textbf{The Philadelphia Eagles won the championship} game, defeating the Dallas Cowboys, 34–3.
 Super Bowl LII was the second title game in which both teams had never won a championship, the first was Super Bowl XXXII, which
\\
\\
Speculative Decode\\
------------------\\
Time = 11.24s\\
Text = \textbf{The Philadelphia Eagles won the championship} in the American Football League ( few years later the AFL merged with the NFL to a new league called the NFL Europe) in 1974.
 the Eagles were coached by Dick Verm had a record of 6 wins and
\\
-----------------------------------------------------------------------\\

\subsubsection{N=100}

Autoregressive Decode\\
---------------------\\
Time = 44.25s\\
Text = \textbf{Explain the difference between fission and fusion} energy
 nuclear fission and nuclear fusion
Fission
Fission is the process by which an atom splits into two or more smaller atoms. Fission typically occurs when a heavy atom (such as uranium) is hit by a neutron. The atom absorbs the neutron and becomes unstable, splitting into two or more smaller atoms with less mass and energy.
Fission reactions release a tremendous amount of energy in the form of gamma rays and other high-energy particles. Because of this,
\\
\\
Speculative Decode\\
------------------\\
Time = 26.97s\\
Text = \textbf{Explain the difference between fission and fusion} energy.
a ITERM reactor is a fusion reactor. It is a huge magnetic field generated by a huge number of electromagnets that are powered by a small amount of fusion fuel
had a huge magnetic field generated by a huge number of electors that are powered by a single amount of fusion fuel.
first started operating in 1954. The ITER project is a collaboration between Europe, China, and the United States
ITER is expected to cost \$,
\\
-----------------------------------------------------------------------------\\
\\
Autoregressive Decode\\
---------------------\\
Time = 43.39s\\
Text = \textbf{Who was better between the Beatles and the Rolling Stones?}
 answer: The Beatles
The Beatles and the Rolling Stones are both very influential bands that had a huge impact on the music industry and the culture as a whole. Many people have strong opinions on which band was better, but it is impossible to say who was "better" because everyone's tastes and values are different.
: The Beatles and the Rolling Stones are both very influential bands that had a huge impact on the music industry and the culture as a whole. Many people have strong
\\
\\
Speculative Decode\\
------------------\\
Time = 30.07s\\
Text = \textbf{Who was better between the Beatles and the Rolling Stones?} did the Beatles have a better chance of winning the war?
first of all, the war was fought in the 1960's and the Beatles were formed in Liverpool in 1960
, so the Beatles had a lot longer to. fight the war than the Germans had to fight the war on the UK
The UK had a much smaller share of the world's population.
, the Beatles had a better album sales record than the Rolling St have had albums sold.
 "afety in numbers "
\\
-----------------------------------------------------------------------------\\
\\
Autoregressive Decode\\
---------------------\\
Time = 41.39s\\
Text = \textbf{I can't seem to sleep lately because} of the noise." "I'm going to the city tomorrow to look for a new house." "You can stay here until you find one." "I'm not going to any city." "Why not?" "I'm going to look for my son." "Your son?" "Do you have a picture of him?" "Yes." "Would you like to see it?" "Yes." "Here." "He's very handsome." "He looks just like you." "Thank you." "
\\
\\
Speculative Decode\\
------------------\\
Time = 24.38s
Text = \textbf{I can't seem to sleep lately because} I'm worried about you." "I'm worried about you, too " " I'm worried about you, too " " I'm worried about you, too " " I'm worried about you, too " " I mean, you're so close to me, and I can't seem to reach you " " I mean I can't seem to reach the shore " " I mean, you're so close to the shore " " I can't seem to reach you "in my heart, you're not
\\
-----------------------------------------------------------------\\
\\
Autoregressive Decode\\
---------------------\\
Time = 41.33s\\
Text = \textbf{I used to live in Los Angeles}, and I used to go to the beach all the time. I used to go to Venice Beach all the time. I used to walk by this place called the Magic Mountain. It's a magic shop, and it's like a little museum. It's got all these old-timey magic tricks and stuff. It's really cool.
 ooOOooohhhh...
I used to live in Los Angeles, and I used to go to the beach all the time. I\\
\\
\\
Speculative Decode\\
------------------\\
Time = 24.62s\\
Text = \textbf{I used to live in Los Angeles}, and I used to go to the beach a lot, and I would always bring him with me I would take him to the airport and he would sit in the lap of the person who was taking me to the airport." "And I would tell him, "Don't be silly, you're not going to the airport in your diaper " have you thought about changing?" " now, now, now, now, now, now, now, now, now, now, now, now, now\\
\\
------------------------------------------------------------------\\
\end{document}